\documentclass[letterpaper, 10 pt, conference]{ieeeconf}

\usepackage{multirow}
\usepackage[cmex10]{amsmath} \interdisplaylinepenalty=2500
\usepackage[most]{tcolorbox}
\usepackage{blkarray}
\usepackage{dsfont}
\usepackage{bm}
\usepackage{comment}
\usepackage{amssymb}
\usepackage{gensymb}
\usepackage{booktabs,makecell}
\usepackage{epsfig,floatflt}
\usepackage{psfrag,graphicx}
\usepackage{xcolor}
\usepackage[colorlinks,allcolors=black]{hyperref}
\usepackage{cite}
\usepackage{epstopdf}
\usepackage{subcaption}
\usepackage[labelsep=quad,skip=3pt]{caption}
\usepackage{tikz}
\usetikzlibrary{arrows.meta,calc,fit,backgrounds,positioning}
\usepackage{cuted}
\usepackage{capt-of}
\newsavebox\rgbox

\IEEEoverridecommandlockouts
\newtheorem{rem}{\textbf{Remark}}

\DeclareSymbolFont{extraup}{U}{zavm}{m}{n}
\DeclareMathSymbol{\vardiamond}{\mathalpha}{extraup}{87}

\begin{document}
	\title{Continuous-Time Safe Planning and Control for Agile Quadrupeds

    \thanks{\textsuperscript{1} Amazon Fulfillment Technologies \& Robotics, Westborough MA, USA. \texttt{\{camtab, jmigdal, bradytye, ptaskin\}@amazon.com}.}
    \thanks{\raggedright\textsuperscript{2} University of Iowa, Iowa City, IA 52240. \texttt{\{gage-maclin, venanzio-cichella\}@uiowa.edu}.}
\thanks{\textsuperscript{3} Massachusetts Institute of Technology, Cambridge, MA 02139, USA. \texttt{masonbp@mit.edu}.}

\thanks{This work was done while Gage MacLin and Mason Peterson were interns at Amazon.}
\thanks{Taşkın Padır holds concurrent appointments as a Professor of Electrical and Computer Engineering at Northeastern University and as an Amazon Scholar.
This paper describes work performed at Amazon and is not associated with Northeastern University.}
}

\author{\authorblockN{Gage MacLin\textsuperscript{1,2}, Mason Peterson\textsuperscript{1,3}, Camilla Tabasso\textsuperscript{1}, Venanzio Cichella\textsuperscript{2}, Joshua Migdal\textsuperscript{1}, Tye Brady\textsuperscript{1}, Taşkın Padır\textsuperscript{1}}}
	\maketitle
	\begin{abstract} Agile quadruped navigation in dynamic environments demands trajectories that are both dynamically feasible and provably collision-free. Most real-time planners achieve tractability through a convex reduction that requires simplified kinematic models, discarding motion that the robot can readily execute, and certify safety only at discretization points, leaving open possible inter-sample collisions with moving obstacles. We present a real-time motion planning framework that retains the fully-actuated $SE(2)$ model of a quadruped, with its anisotropic, heading-coupled velocity bounds, and accepts the resulting nonconvexity. By parameterizing trajectories as composite Bernstein polynomials, we exploit the convex hull property to certify collision avoidance and the input bounds over the entirety of each planning horizon. Solved at online replanning rates, the optimization runs in closed loop, commanding body velocities to the locomotion controller. We validate the approach in simulation against ablations and external baselines, and in hardware experiments on a quadruped navigating among unknown moving obstacles, recovering both lateral and heading-coupled motion that simplified models exclude.
	\end{abstract}
	

	\IEEEpeerreviewmaketitle


\section{Introduction}\label{sec:Introduction}

Quadrupedal robots are transitioning from demonstration platforms to working tools on construction sites, in warehouses, and in other spaces shared with people, vehicles, and machines. Advances in model predictive control (MPC) and whole-body control have enabled dynamic gaits, omnidirectional locomotion, and traversal of unstructured terrain \cite{di2018dynamic, grandia2023perceptive}. In particular, quadrupeds admit a rich set of motion primitives, including lateral translation and heading-dependent velocity limits, i.e., distinct bounds on longitudinal and lateral body velocities. Yet, to remain computationally tractable, the high-level planners that provide references for these controllers typically plan over simplified kinematic models, discarding much of this motion capability \cite{molnar2023safety, lin2025one, dudzik2020robust}.

\begin{figure}[t]\centering
  \includegraphics[width=\linewidth]{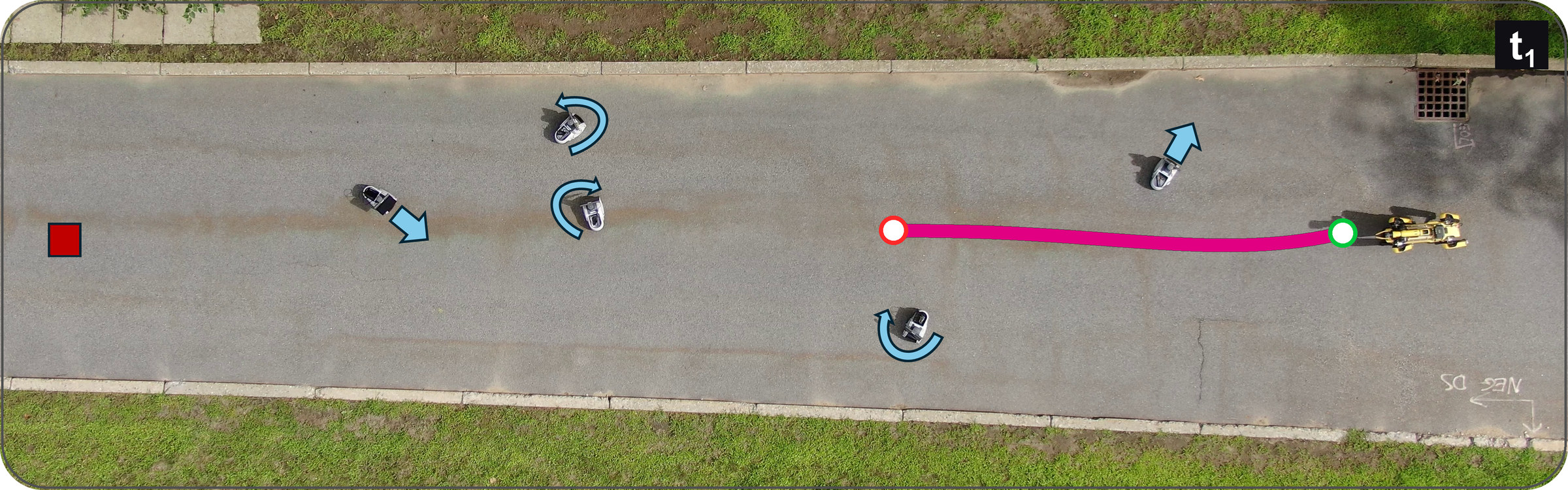}\\[3pt]
  \includegraphics[width=\linewidth]{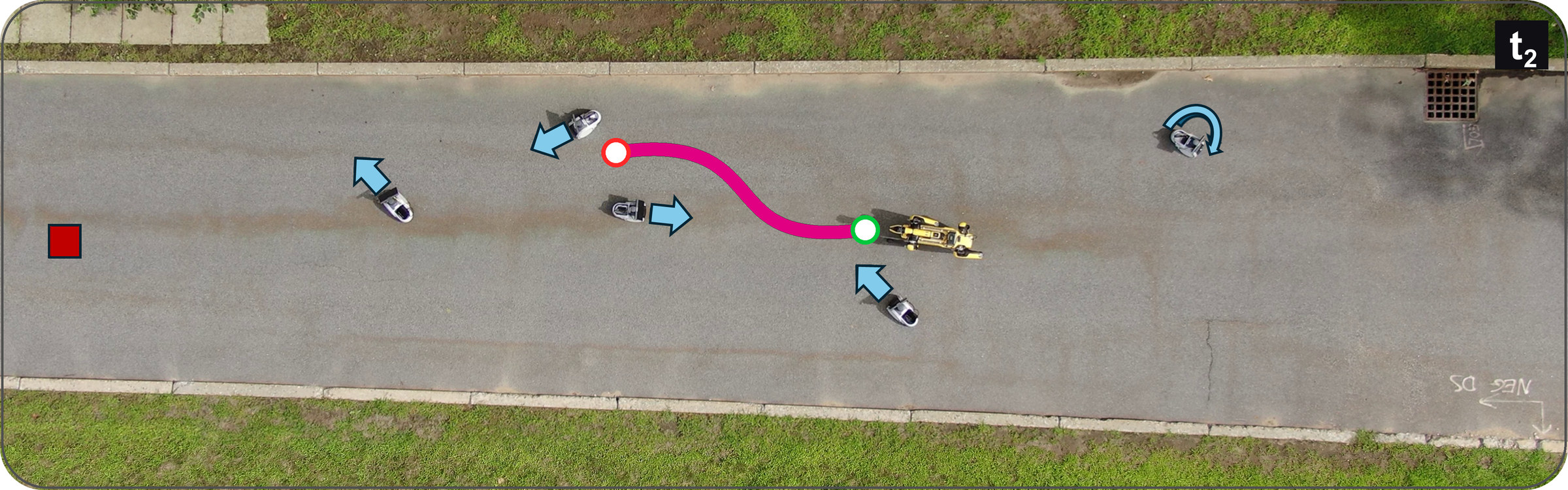}\\[3pt]
  \includegraphics[width=\linewidth]{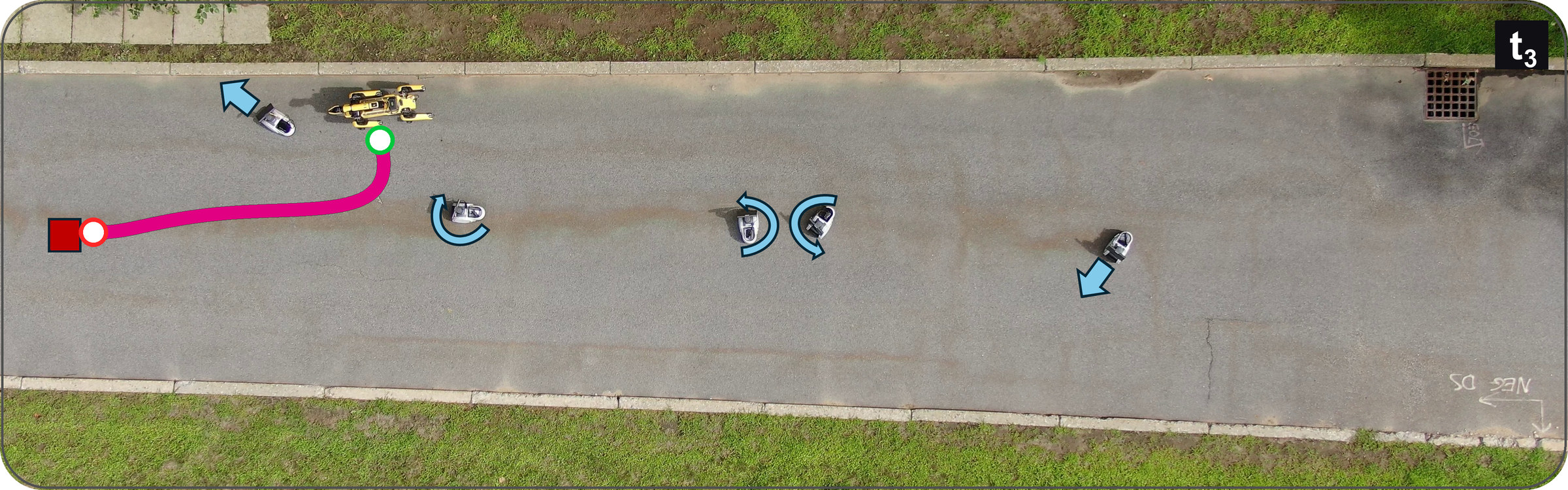}
\caption{Spot crossing a field of five differential-drive robots using our planner, shown at timesteps $t_1$--$t_3$ as obstacles cross its path. The magenta curve is the planned trajectory, arrows indicate obstacle motion.}
\label{fig:hardware}
\end{figure}


This conservatism has computational origins, inherited from aerial robotics, where real-time trajectory optimization first matured. Quadrotor motion can be accurately approximated via differentially flat kinematic models, where the full state and input are recovered algebraically from a low-dimensional flat output and its derivatives. Dynamic feasibility can then be enforced by construction on a polynomial trajectory, often allowing planning to reduce to a convex program rather than a traditionally burdensome nonlinear program (NLP) \cite{mellinger2011minimum, wang2022geometrically}. Preserving the convex program requires convexifying the collision constraints as well, commonly by confining the trajectory to safe corridors of convex polytopes \cite{kondo2026sando}. Other planners forgo convexity in the obstacle terms altogether, introducing separating-hyperplane decision variables \cite{tordesillas2021mader} or soft gradient penalties on obstacle proximity \cite{zhou2020ego}. Combined with polynomial bases such as B\'ezier and B-spline, whose control points bound the trajectory and its derivatives \cite{tordesillas2022minvo}, flatness is used to enable replanning at high rates even in cluttered and dynamic environments \cite{tordesillas2021mader}.

For a quadruped, however, this pipeline breaks at the velocity limits. A quadruped's velocity commands are bounded in the body frame, with distinct longitudinal and lateral limits that rotate with the heading, and these bounds are nonconvex in the flat outputs, so the flat structure yields no computational advantage. A planner that insists on the convex reduction must then downgrade the model, often to an integrator that ignores the heading coupling or to a unicycle that forbids lateral motion, and in either case it discards motion that the locomotion controller can readily execute.

Our formulation abandons the convex reduction in two ways. The kinematics enter as nonlinear equality constraints, and collision avoidance enters as hard keep-out constraints that are nonconvex in the decision variables. This cost is no longer prohibitive, as nonlinear planners can run at rates compatible with online replanning for ground robots \cite{rosmann2015timed}, quadruped navigation \cite{zhang2024agile}, and non-differentially flat systems \cite{dirckx2026accelerated}, aided by structure-exploiting transcriptions, mature sparse NLP solvers, and warm-started replanning. 

A conventional direct transcription, which discretizes the trajectory and enforces the constraints at the resulting nodes, leaves the trajectory uncertified between them. We instead parameterize the pose and input trajectories as composite Bernstein polynomials~\cite{cichella2020optimal,maclin2024optimal}, sequences of low-degree Bernstein polynomials joined at knots. A polynomial in this basis lies within the convex hull of its coefficients, so bounding the coefficients bounds the continuous curve, a property that flat planners already exploit for feasibility~\cite{tordesillas2021mader, morando2025trajectory}, and one that is least conservative when the degree is low. The velocity bounds and the obstacle clearance are polynomial functions of the trajectory, so the convex hull property can be used to certify these constraints for all $t$.

We show that the computational cost of this nonconvexity is modest, and that the recovered lateral and heading-coupled motion allows a quadruped to reach goals faster and with fewer collisions than one planned with a simplified model. Our contributions are as follows:
\begin{itemize}
    \item We formulate quadruped trajectory generation as a receding-horizon NLP, replanning as the robot moves, over composite Bernstein polynomials using the fully-actuated $SE(2)$ kinematics, with anisotropic velocity and collision constraints certified for all $t$.
    \item We show that the resulting NLP solves in 16\,ms median on an onboard NVIDIA Jetson, replanning at $\sim36$\,Hz, twice the refresh rate of the onboard perception.
    \item We demonstrate in closed-loop simulations over 50 scenarios that our planner reaches the goal 1.6--2.7$\times$ faster than every tested kinematic ablation and external baseline, with the fewest collisions among moving obstacles.
    \item We deploy the full stack (Fig.~\ref{fig:overview}) onboard a real quadruped across 22 consecutive traversals totaling 392\,m among unknown moving robots, with perception, planning, and control running in real time.
\end{itemize}


\section{Related Works}
\subsection{Quadruped Locomotion and Navigation}
Convex MPC and whole-body control have made dynamic quadrupedal locomotion reliable enough to treat the low-level controller as a velocity-tracking abstraction \cite{di2018dynamic, grandia2023perceptive}. Safety-critical extensions embed obstacle avoidance directly into the locomotion controller, through soft collision constraints \cite{gaertner2021collision} or discrete control barrier functions with duality-based polytope avoidance \cite{liao2023walking}. Both enforce safety at discrete controller nodes, and require access to a whole-body control layer that commercial platforms often do not expose, whereas our planner certifies safety continuously at the planning layer through the velocity interface. Navigation frameworks built on this abstraction, however, typically decompose the problem hierarchically, pairing a geometric or sampling-based global planner with a local trajectory optimizer over a simplified model \cite{dudzik2020robust}. Motion anisotropy has likewise been addressed with an elliptical velocity constraint, imposed within the hierarchical pipeline where a LazyPRM*-based kinodynamic search seeds a nonlinear trajectory optimization \cite{zhang2024agile}. We instead solve a single NLP whose velocity and collision constraints are certified over the continuous trajectory (Sec.~\ref{sec:method}).

\subsection{Flatness-Based Real-Time Planning}
For differentially flat systems, dynamic feasibility is enforced by construction on a polynomial flat output, reducing trajectory optimization to a convex program \cite{mellinger2011minimum} and to lightweight unconstrained problems in spatial-temporal formulations \cite{wang2022geometrically}, with bases such as B\'ezier and MINVO providing beneficial geometric properties \cite{tordesillas2022minvo}. Because obstacle keep-out constraints are often nonconvex, these planners handle them separately, through safe corridors of polytopes \cite{kondo2026sando}, graphs of convex sets \cite{marcucci2023motion}, separating hyperplanes \cite{tordesillas2021mader}, or soft gradient costs \cite{zhou2020ego}. These strategies recover tractability, though corridors conservatively approximate free space and soft penalties provide no strict clearance guarantee. Differential flatness of the car-like model enables real-time spatial-temporal trajectory optimization for autonomous vehicles in unstructured environments \cite{han2023efficient}, and fixed-wing aircraft in coordinated flight admit flatness-based Bernstein planning with online replanning \cite{morando2025trajectory}. These models achieve impressive replanning rates, but only for flat models whose constraints remain convex in the flat output. We retain the nonconvex $SE(2)$ model to recover motion primitives that simpler models discard, and we enforce the clearance inequalities as hard polynomial constraints along the continuous trajectory, accepting a bounded disc-footprint conservatism rather than the environment-dependent conservatism of corridors.

\subsection{Bernstein-Based Nonlinear Trajectory Optimization}
Direct methods transcribe optimal control problems into NLPs, and the Bernstein basis suits this transcription because its convex hull, endpoint, and closure properties (Sec.~\ref{sec:bernstein}) allow path constraints to be enforced along the whole trajectory rather than only at sampled nodes \cite{cichella2020optimal}. Composite Bernstein polynomials~\cite{maclin2024optimal} converge faster than a single polynomial and capture sharp local features that a global polynomial smooths over. The same hull property yields continuous-time certificates against polytope obstacles for non-flat systems in~\cite{dirckx2026accelerated}, where separation is certified by hyperplanes computed between solver iterations. Whether such certificates exist at all depends on the chosen footprint and obstacle representations, since the clearance function must remain polynomial in the trajectory for continuous-time guarantees. With obstacles approximated as circles and a disc footprint for the robot (Fig.~\ref{fig:spot}), our clearance constraints are polynomial, with an exact certificate for every executed heading implicitly, and no additional separation variables. The drawback is excess clearance, at most the lateral gap between the body and the circumscribing circle. Polytope footprints can be  less conservative, but a polytope frozen at the solve's initial heading is certified for that heading only, and orientation-tracking polytopes, e.g., via duality~\cite{liao2023walking}, are not polynomial, forfeiting continuous-time certificates.

\subsection{Learning-Based Approaches}
Learning-based methods have been implemented across the navigation stack, from reinforcement-learning locomotion controllers \cite{hwangbo2019learning} to end-to-end policies that command velocities directly from perception \cite{hoeller2024anymal}, absorbing the planner into the policy. Such policies react directly to raw sensing, with no explicit obstacle model or geometric pipeline in between, and extend to terrain that resists clean geometric representation. Their collision avoidance, however, is encouraged by training rather than guaranteed by construction. Reliable avoidance is therefore often obtained by pairing the policy with a model-based safety filter \cite{hsu2023safety}, which returns the problem to the constrained-optimization setting that we adopt outright, where hard constraints hold by construction, and behavior is shaped through the cost and constraint set, with no training.

\section{Problem Formulation}
\subsection{Fully-Actuated $SE(2)$ Model}
\label{sec:model}
We consider a quadruped whose locomotion controller accepts body-frame velocity commands, and we plan over the 2D special Euclidean group $SE(2)$
\begin{equation}
\label{eq:planar}
\underbrace{\begin{bmatrix} \dot{x} \\ \dot{y} \\ \dot{\theta} \end{bmatrix}}_{\dot{\mathbf{q}}}
=
\underbrace{\begin{bmatrix}
\cos\theta & -\sin\theta & 0 \\
\sin\theta & \phantom{-}\cos\theta & 0 \\
0 & 0 & 1
\end{bmatrix}}_{\bm{R}(\theta)}
\underbrace{\begin{bmatrix} v_x \\ v_y \\ \omega \end{bmatrix}}_{\mathbf{u}},
\end{equation}
where $\mathbf{q} = [x, y, \theta]^\top \in SE(2)$ is the body pose in the world frame, and the input $\mathbf{u} = [v_x, v_y, \omega]^\top$ collects the longitudinal velocity $v_x$, lateral velocity $v_y$, and yaw rate $\omega$, all in the body frame. Unlike the unicycle model, which is recovered by fixing $v_y \equiv 0$, the fully-actuated $SE(2)$ model retains an independent lateral velocity and heading.

\begin{rem}[Flatness does not yield a convex program]
\label{rem:flatness}
The model \eqref{eq:planar} is differentially flat in the trivial sense. Taking the pose itself as the flat output $\mathbf{y} = \mathbf{q}$, the inputs are recovered algebraically as
\begin{equation}
\label{eq:flatmap}
v_x = \dot{x}\cos\theta + \dot{y}\sin\theta, \quad
v_y = -\dot{x}\sin\theta + \dot{y}\cos\theta, \quad
\omega = \dot{\theta}.
\end{equation}
For flat systems, such as those used for quadrotors, this structure reduces trajectory optimization to a convex program because the relevant feasibility constraints are convex functions of the flat output and its derivatives. Here that reduction fails, substituting \eqref{eq:flatmap} into the velocity bounds of Sec.~\ref{sec:constraints}, e.g., the elliptical bound \eqref{eq:ellipse}, yields
\begin{equation}
\label{eq:flatconstraint}
\frac{\left(\dot{x}\cos\theta + \dot{y}\sin\theta\right)^2}{v_{x,\max}^2}
+ \frac{\left(-\dot{x}\sin\theta + \dot{y}\cos\theta\right)^2}{v_{y,\max}^2} \le 1,
\end{equation}
which is trigonometric in $\theta$ and bilinear between the translational rates and functions of the heading, hence nonconvex in the flat outputs. The flat parameterization therefore offers limited computational advantage, and we formulate the problem directly on \eqref{eq:planar}, accepting the resulting NLP.
\end{rem}

\begin{figure}[t]\centering
\vspace{4pt}
\resizebox{\columnwidth}{!}{\input{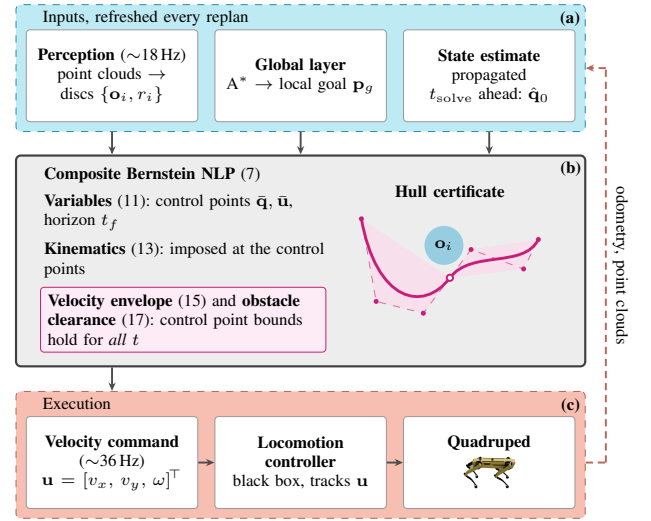}}
\caption{Planning pipeline. Each replan (a) refreshes the obstacle set, goal, and propagated initial state, (b) solves the composite Bernstein NLP~\eqref{eq:ocp}, certifying the velocity and obstacle-clearance constraints for all $t$ via the control-point hull, and (c) tracks the body-velocity command, with odometry and point clouds fed back to (a).}
\label{fig:overview}
\end{figure}

\subsection{Anisotropic Motion Constraints}
\label{sec:constraints}
The quadruped's velocity capability is anisotropic, with the achievable longitudinal speed exceeding the lateral, as the two are coupled through the legs' shared ground-reaction forces \cite{zhang2024agile}. We capture this envelope as the intersection of an elliptical bound and box bounds in the body frame (Fig.~\ref{fig:spot}a),

\begin{equation}\label{eq:ellipse}
   \frac{v_x^2}{v_{x,\max}^2} + \frac{v_y^2}{v_{y,\max}^2} \le 1,
\end{equation}

\begin{equation}\label{eq:box}
    -v_{x,\mathrm{rev}} \le v_x \le v_{x,\max},\quad|v_y| \le v_{y,\max}, \quad |\omega| \le \omega_{\max},
\end{equation}
where $v_{x,\max} > v_{y,\max} > 0$ are the maximum longitudinal and lateral speeds, $v_{x,\mathrm{rev}} < v_{x,\max}$ is the maximum reverse speed, and $\omega_{\max} > 0$ is the maximum yaw rate.

With \eqref{eq:ellipse} and \eqref{eq:box} posed in the body frame, they couple the feasible translational velocity to the heading $\theta$ through \eqref{eq:planar}. This coupling is the source of the nonconvexity identified in Remark~\ref{rem:flatness} and it is what simplified models discard. Note that a single-integrator model replaces \eqref{eq:ellipse} with a heading-independent bound, overestimating the feasible velocity set in the lateral direction, while the unicycle enforces $v_y \equiv 0$, underestimating it.

\subsection{Collision Avoidance}
\label{sec:collision}
\begin{figure}[t]
    \centering
    \includegraphics[width=\columnwidth]{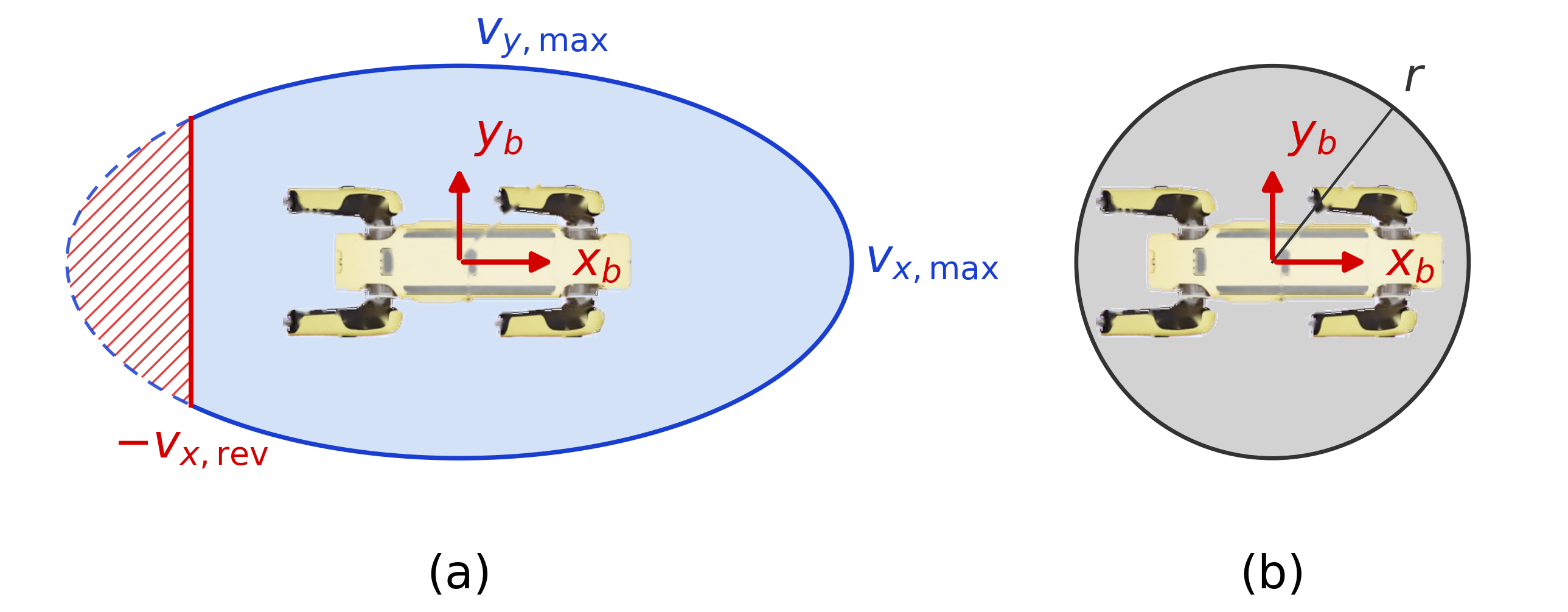}
    \caption{(a) anisotropic velocity bounds, (b) disc footprint.}
    \label{fig:spot}
\end{figure}

The robot footprint is a disc of radius $r_{\mathrm{rob}}$ centered at the body position $\mathbf{p}(t) = [x(t), y(t)]^\top$, circumscribing the body (Fig.~\ref{fig:spot}b). Its conservatism is bounded, adding approximately $0.05$\,m longitudinally and $0.35$\,m laterally, with certified clearance of the disc implying clearance of the enclosed body.

The perception pipeline (Sec.~\ref{sec:system}) supplies $N_o$ circular obstacles, the $i$th with center $\mathbf{o}_i \in \mathbb{R}^2$ and radius $r_i$. Collision avoidance requires
\begin{equation}
\label{eq:collision}
\left\lVert \mathbf{p}(t) - \mathbf{o}_i \right\rVert^2 \ge \left( r_i + r_{\mathrm{rob}} + \epsilon \right)^2, \quad \forall t, \; i = 1, \dots, N_o,
\end{equation}
where $\epsilon > 0$ is a clearance margin. Because the black-box locomotion controller tracks the commanded velocities rather than $\mathbf{p}(t)$ itself, \eqref{eq:collision} certifies the planned trajectory, with the executed path deviating by some tracking error. This deviation is bounded by the error accrued within one replan cycle, and $\epsilon$ is chosen to mitigate it. Each constraint \eqref{eq:collision} is nonconvex, and we impose it as a hard constraint certified over the entire trajectory rather than at sampled instants. Sec.~\ref{sec:method} shows how the composite Bernstein parameterization achieves this. The obstacle set is refreshed as often as possible from perception and treated quasi-statically, with each solve certifying continuous-time clearance with respect to the obstacle positions perceived at that solve. Obstacle motion between solves is uncertified, and its effect shrinks with the replanning rate. In this work, we do not assume obstacle trajectory predictions are available. However, when they are, the formulation extends directly, since for polynomial predictions $\mathbf{o}_i(t)$ the clearance function \eqref{eq:collision} remains a polynomial in $t$, and the same hull certification applies along the predicted motion.

\subsection{Receding-Horizon Trajectory Optimization}
\label{sec:ocp}
Each solve takes $t_{\text{solve}}$ to complete, during which the robot keeps moving under its planned command. The planner therefore plans not from the current state estimate but from the pose $\hat{\mathbf{q}}_0$ predicted $t_{\text{solve}}$ ahead in time, obtained by forward-propagating the estimate through \eqref{eq:planar}, so the new trajectory is consistent with the robot's motion at the instant it is deployed. The planner operates in a receding-horizon fashion over a window of bounded free length $t_f \in [t_{\min}, t_{\max}]$. Each solve plans over $t \in [t_0, t_0 + t_f]$ toward the current intermediate goal $\mathbf{p}_g = [x_g, y_g]^\top$ supplied by an $A^*$ global planner (Sec.~\ref{sec:exp-setup}). The trajectory generation problem is
\begin{subequations}\label{eq:ocp}
\begin{flalign}
&\qquad\,\,\min_{\mathbf{q}(\cdot),\,\mathbf{u}(\cdot),\,t_f}
J\big(\mathbf{q}(\cdot), \mathbf{u}(\cdot), t_f\big) && \label{eq:ocp-cost}\\
&\;\;\text{s.t.}\;\;
\dot{\mathbf{q}}(t) = \bm{R}\big(\theta(t)\big)\mathbf{u}(t),
&& t \in [t_0, t_0+t_f], \label{eq:ocp-dyn}\\
&\phantom{\;\;\text{s.t.}\;\;}
\eqref{eq:ellipse}, \eqref{eq:box},
&& t \in [t_0, t_0+t_f], \label{eq:ocp-vel}\\
&\phantom{\;\;\text{s.t.}\;\;}
\eqref{eq:collision},
&& t \in [t_0, t_0+t_f], \label{eq:ocp-col}\\
&\phantom{\;\;\text{s.t.}\;\;}
\mathbf{q}(t_0) = \hat{\mathbf{q}}_0, \quad \mathbf{u}(t_0) = \mathbf{u}_0.
&& \label{eq:ocp-init}
\end{flalign}
\end{subequations}
with $J$ selected for soft terminal attraction to the goal position plus a yaw-rate smoothing term as
\begin{equation}
\label{eq:cost}
J = W_p \left\lVert \mathbf{p}_f - \mathbf{p}_g \right\rVert^2
  + W_\omega \int_{t_0}^{t_0+t_f} \omega^2 \, dt,
\end{equation}
where $W_p$ and $W_\omega$ are weights, and $\mathbf{p}_f = \mathbf{p}(t_0 + t_f)$. Within the fixed window the terminal attraction drives maximum feasible progress, so the closed loop moves at maximum envelope velocity towards each goal. With no time or horizon length terms, solutions typically ride the upper bound $t_{\max}$ until the robot is in tight configurations or approaches the goal position. Mission-level objectives are delegated to the global guidance layer, with success rates reported in Sec.~\ref{sec:experiments}, while safety is the property certified per solve. Because the inputs are velocities, a trajectory that halts and holds is always feasible, so the planner never trades safety for progress. Time-parameterized planners instead either optimize segment durations~\cite{zhang2024agile, wang2022geometrically, dirckx2026accelerated} or fix them by heuristic and re-solve when infeasible~\cite{tordesillas2021mader}. Our bounded window sits in between these positions, and is approximately aligned with the perception range.

\section{Composite Bernstein Trajectory Optimization}
\label{sec:method}
This section transcribes the OCP \eqref{eq:ocp} into a finite-dimensional NLP. The problem is nonconvex in three ways, the $SE(2)$ kinematics \eqref{eq:ocp-dyn}, the heading-coupled velocity constraints \eqref{eq:ocp-vel} and the obstacle avoidance constraints \eqref{eq:ocp-col}. Reducing its trajectories to finitely many decision variables is where a conventional direct transcription \cite{betts1998survey} forfeits the continuous-time guarantee. We make the discretization with composite Bernstein polynomials, parameterizing the pose and input trajectories (Sec.~\ref{sec:param}), imposing the kinematics \eqref{eq:ocp-dyn} at the coefficients (Sec.~\ref{sec:dynamics}), and certifying the velocity and collision constraints along the entire continuous trajectory through the convex hull property (Sec.~\ref{sec:certification}).

\subsection{Bernstein Preliminaries}
\label{sec:bernstein}
The Bernstein basis of degree $N$ on the interval $I_k = [t_{k-1}, t_k]$ is
\begin{equation}
\label{eq:basis}
b_{j,N}^{[k]}(t) = \binom{N}{j} \frac{(t - t_{k-1})^j (t_k - t)^{N-j}}{(t_k - t_{k-1})^N}, \quad j = 0, \dots, N,
\end{equation}
and a Bernstein polynomial on $I_k$ is defined as 
\begin{equation}
    p^{[k]}(t) = \sum_{j=0}^{N} \bar{p}_j^{[k]}\, b_{j,N}^{[k]}(t)
\end{equation} 
where $\bar{p}_j^{[k]}$ are the coefficients (referred to hereafter as control points). Given knots $t_0 < t_1 < \dots < t_K$, a composite Bernstein polynomial is the piecewise function equal to $p^{[k]}(t)$ on each $I_k$. We use three properties of this approximant throughout:
\begin{enumerate}
    \item[(P1)] \emph{Convex hull:} $\min_j \bar{p}_j^{[k]} \le p^{[k]}(t) \le \max_j \bar{p}_j^{[k]}$ for all $t \in I_k$. Bounds on the control points certify bounds on the polynomial over the whole interval.
    \item[(P2)] \emph{Endpoint interpolation:} $p^{[k]}(t_{k-1})$ = $\bar{p}_0^{[k]}$ and $p^{[k]}(t_k)$ = $\bar{p}_N^{[k]}$. Boundary and knot-continuity conditions are linear constraints on the first and last control points.
    \item[(P3)] \emph{Closure under arithmetic and calculus:} sums, products, derivatives and integrals of Bernstein polynomials are again Bernstein polynomials whose control points are linear (sums, derivatives, integrals) or bilinear (products) functions of the original control points, computable through fixed sparse matrices \cite{cichella2020optimal}.
\end{enumerate}
Property (P1) guarantees path constraint satisfaction along the entire trajectory, not only at the discretization points. Property (P3) extends this to any constraint that is a polynomial function of the trajectory, which covers the velocity envelope \eqref{eq:ellipse}--\eqref{eq:box} and the collision constraints \eqref{eq:collision}. Property (P2) enforces boundary conditions by construction.

\subsection{Trajectory Parameterization}
\label{sec:param}
We parameterize the pose and input trajectories as composite Bernstein polynomials of degree $N$ over $K$ segments,
\begin{equation}
\label{eq:param}
\mathbf{q}(t) = \sum_{j=0}^{N} \bar{\mathbf{q}}_j^{[k]}\, b_{j,N}^{[k]}(t), \quad
\mathbf{u}(t) = \sum_{j=0}^{N} \bar{\mathbf{u}}_j^{[k]}\, b_{j,N}^{[k]}(t), \quad t \in I_k,
\end{equation}
with the control points $\bar{\mathbf{q}}_j^{[k]} \in \mathbb{R}^3$, $\bar{\mathbf{u}}_j^{[k]} \in \mathbb{R}^3$ for $j = 0, \dots, N$, $k = 1, \dots, K$, and the bounded horizon $t_f \in [t_{\min}, t_{\max}]$ of \eqref{eq:ocp}. The knots are uniformly spaced, $t_k = t_0 + k\,h$ with segment duration $h = t_f/K$, so time is one scalar decision variable rather than a free knot vector. This sacrifices adaptive knot placement, which is beneficial for problems with isolated discontinuities, in exchange for fewer variables in the receding-horizon setting where trajectories are smooth. Continuity of the composite trajectory at the interior knots is imposed through (P2) as the linear constraints
\begin{equation}
\label{eq:continuity}
\bar{\mathbf{q}}_N^{[k]} = \bar{\mathbf{q}}_0^{[k+1]}, \qquad
\bar{\mathbf{u}}_N^{[k]} = \bar{\mathbf{u}}_0^{[k+1]}, \qquad k = 1, \dots, K-1,
\end{equation}
which enforce $C^0$ continuity of the pose and of the input. With the inputs being the commanded body velocities, input continuity guarantees that the velocity commands delivered to the locomotion controller contain no jumps. The boundary conditions \eqref{eq:ocp-init} become linear constraints on the first control points of segment $1$. We note that heading references and constraints, such as pinning $\theta$ or tracking an inspection target, are linear constraints on the $\theta$ control points, supporting heading-constrained tasks directly, although we do not evaluate them here.

\subsection{Enforcement of the Kinematics}
\label{sec:dynamics}
Because of the rotation matrix, the kinematics \eqref{eq:ocp-dyn} are not polynomial unless the heading is constant or the body rotates in place. No polynomial trajectory satisfies them exactly, so some approximation is unavoidable. 
With $\dot{\bar{\mathbf{q}}}_j^{[k]}$ denoted as the control points of $\dot{\mathbf{q}}^{[k]}(t)$ after degree elevation to order $N$ (re-expressing the degree $N$-$1$ derivative in the degree $N$ basis), the kinematics are imposed as the equalities
\begin{equation}
\label{eq:collocation}
\dot{\bar{\mathbf{q}}}_j^{[k]} = \bm{R}\big(\bar{\theta}_j^{[k]}\big)\, \bar{\mathbf{u}}_j^{[k]}, \quad j = 0, \dots, N, \; k = 1, \dots, K,
\end{equation}
matching each derivative control point to the vector field evaluated at the corresponding pose and input control points. Control points coincide with the trajectory only at segment endpoints, so \eqref{eq:collocation} constrains the control points themselves rather than the polynomial evaluated at their node times. The approximation is nonetheless convergent, with feasible solutions of the transcription approaching solutions of the continuous problem at rate $\mathcal{O}(1/K^2N)$ \cite{maclin2024optimal}.
 
How the transcription distributes its approximation is a critical design principle. A poor kinematic approximation reaches the robot as a tracking error, which the locomotion controller can absorb, while a constraint violation between samples reaches it as a collision or an infeasible command. We therefore discretely approximate the kinematics, where its consequences are more benign, and keep the velocity and collision constraints exact along the entire continuous trajectory (Sec.~\ref{sec:certification}). A flat parameterization would allocate the error the other way around, with the pose satisfying the kinematics exactly and the inputs derived from the flat map \eqref{eq:flatmap}, but the velocity constraints then become the non-polynomial expressions of Remark~\ref{rem:flatness} and cannot be certified efficiently. 

\subsection{Continuous Constraint Certification}
\label{sec:certification}
The central benefit of the Bernstein parameterization is that the nonconvex path constraints of Sec.~\ref{sec:ocp} are polynomial in the trajectory and can therefore be certified over the entire horizon through (P1) and (P3).

\subsubsection{Velocity envelope}
The elliptical constraint \eqref{eq:ellipse} evaluated along \eqref{eq:param} defines, on each segment,
\begin{equation}
\label{eq:ellipse-bern}
g^{[k]}(t) = \frac{v_x^{[k]}(t)^2}{v_{x,\max}^2} + \frac{v_y^{[k]}(t)^2}{v_{y,\max}^2} - 1,
\end{equation}
a polynomial of degree $2N$ whose Bernstein control points $\bar{g}_j^{[k]}$, $j = 0, \dots, 2N$, are quadratic functions of the input control points, computed through the fixed product matrices of (P3). By the convex hull property, the constraints
\begin{equation}
\label{eq:ellipse-cert}
\bar{g}_j^{[k]} \le 0, \quad j = 0, \dots, 2N, \; k = 1, \dots, K,
\end{equation}
certify $g^{[k]}(t) \le 0$ for all $t$, i.e., satisfaction of \eqref{eq:ellipse} along the entire continuous trajectory. The box bounds are certified directly by bounding the input control points by $|\bar{v}_{x,j}^{[k]}| \le v_{x,\max}$, $|\bar{v}_{y,j}^{[k]}| \le v_{y,\max}$, $|\bar{\omega}_j^{[k]}| \le \omega_{\max}$, and $\bar{v}_{x,j}^{[k]}\ge-v_{x,\text{rev}}$.

\subsubsection{Collision avoidance}
Similarly, for each obstacle $i$ the clearance function
\begin{equation}
\label{eq:collision-bern}
h_i^{[k]}(t) = \left\lVert \mathbf{p}^{[k]}(t) - \mathbf{o}_i \right\rVert^2 - \left( r_i + r_{\mathrm{rob}} + \epsilon \right)^2
\end{equation}
is a polynomial of degree $2N$ with control points $\bar{h}_{i,j}^{[k]}$ that are quadratic in the position control points, and the constraints
\begin{equation}
\label{eq:collision-cert}
\bar{h}_{i,j}^{[k]} \ge 0, \quad j = 0, \dots, 2N, \; k = 1, \dots, K, \; i = 1, \dots, N_o,
\end{equation}
certify obstacle clearance for all $t \in [t_0, t_0 + t_f]$ with respect to the obstacle set of the current solve (Sec.~\ref{sec:collision}). No safe corridor, sampling grid, or soft penalty is involved, as \eqref{eq:collision-cert} is a hard constraint whose satisfaction implies safety of the continuous trajectory against that set.


\subsection{Resulting NLP and Real-Time Implementation}
\label{sec:nlp}
Collecting \eqref{eq:continuity}, \eqref{eq:collocation}, \eqref{eq:ellipse-cert}, \eqref{eq:collision-cert}, the box bounds, and the boundary conditions, and evaluating the cost \eqref{eq:ocp-cost} on the parameterization \eqref{eq:param} yields an NLP with the $6K(N+1)$ trajectory control points and the single bounded time variable $t_f$ as decision variables. The problem is structured with every constraint involving only the control points of a single segment (and, for \eqref{eq:continuity}, its neighbor), so the constraint Jacobian is block-banded, i.e., each constraint touches only a few variables, and the number of variables is small as $N$ is low, while (P1)-(P3) hold per segment. Each replan is warm-started directly from the previous solution, with the optimal control points and horizon $t_f$ of the preceding solve reused as the initial guess. The first solve is cold-started from a straight-line interpolation between the initial and goal position, and zeros otherwise. The NLP is solved with CasADi using IPOPT. For the simulation and hardware results below, the Bernstein-based planners were run with $K=2$, $N=5$, $t_{\min}=0.1$\,s, $t_{\max}=1$\,s, and $r_\text{rob}=0.6$\,m.

\section{Experiments}\label{sec:experiments}
We evaluate the planner in two stages. A simulation benchmark compares the proposed planner against ablations of its kinematic model within the same framework and against two external baselines on identical seeded scenarios. A hardware deployment then runs the full stack onboard a Boston Dynamics Spot under real sensing.

\subsection{Simulation Setup}\label{sec:exp-setup}
Simulations run in Isaac Sim on an Intel i9-13950HX with an RTX 4090 Laptop GPU, with NVIDIA's \texttt{SpotFlatTerrainPolicy} locomotion controller. The planner is configured as in Sec.~\ref{sec:nlp}, with the bounds set to $v_{x,\max}=3$\,m/s,
$v_{y,\max}=1.5$\,m/s, $\omega_{\max}=2.0$\,rad/s, and
$v_{x,\mathrm{rev}}=1.5$\,m/s, with $W_p=1.0$ and $W_{\omega}=5.0$.

\begin{figure}[t]
\vspace{4pt}
    \centering
    \includegraphics[width=\columnwidth]{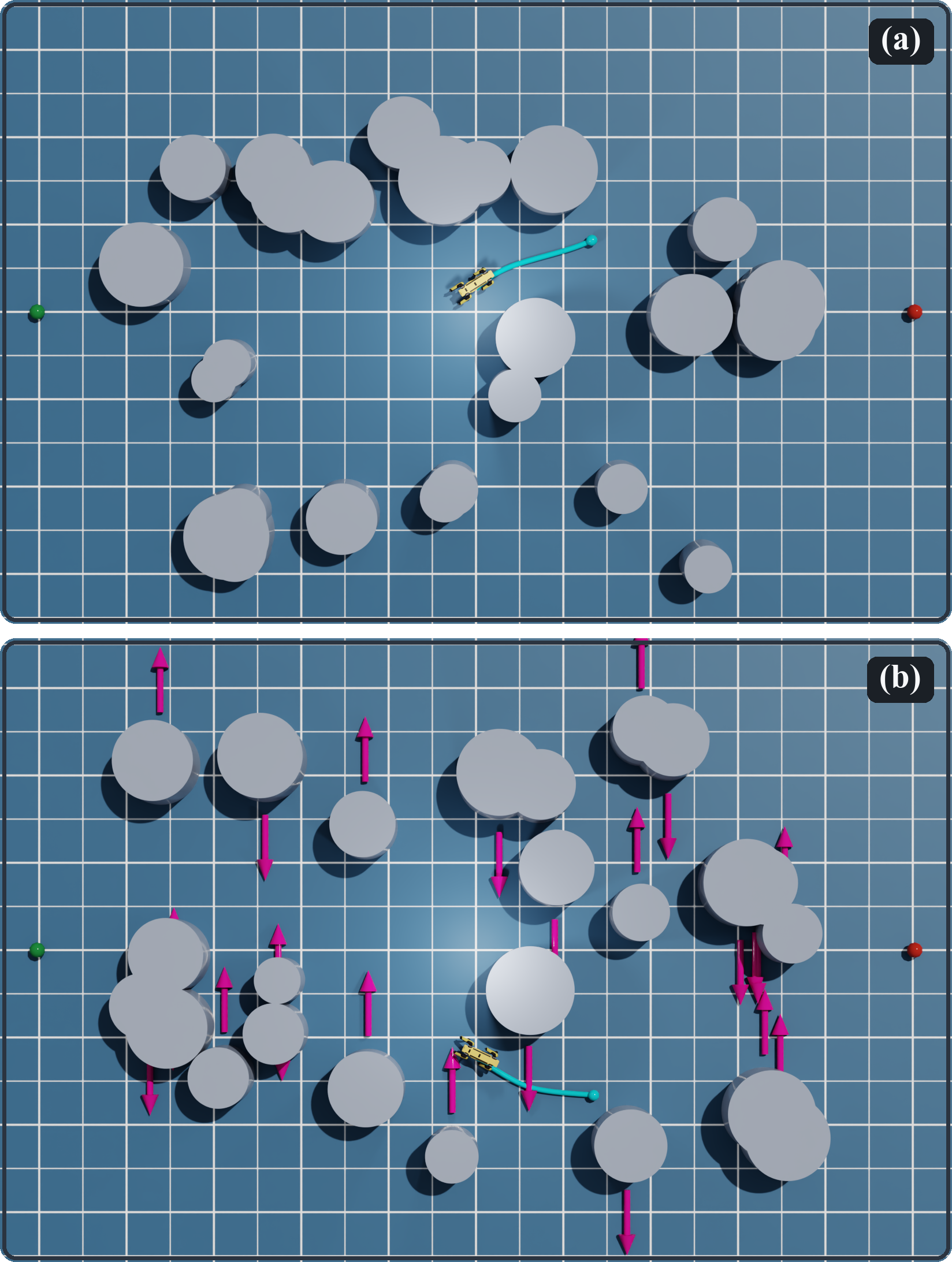}
    \caption{Benchmark environments, rendered in Isaac Sim.
    (a)~Static condition: the 25 cylindrical obstacles are held
    fixed. (b)~Dynamic condition: identical seeded obstacle field,
    each obstacle crosses laterally back-and-forth in its own lane.}
    \label{fig:environments}
\end{figure}

The task is a 20\,m traversal through a field of 25 cylindrical obstacles of radius $r\in[0.5,1.0]$\,m, procedurally generated from set seeds so that every planner faces an identical layout on a given seed (Fig.~\ref{fig:environments}). A \emph{static} condition holds the obstacles fixed. A \emph{dynamic} condition uses the same fields but each obstacle crosses the corridor laterally, shuffling back and forth in its own lane at a constant speed in $[0.5,1.0]$\,m/s, with perceived positions updating at 15\,Hz. The dynamic condition stresses the quasi-static obstacle treatment of Sec.~\ref{sec:collision}, where plans certified against a snapshot must survive obstacle motion between replans.

Each trial ends in one of three mutually exclusive outcomes. A collision is when Spot's articulated body comes into contact with any obstacle, detected from PhysX contact forces. A collision-free trial is a success if the body reaches within 1.0\,m of the goal before a 40\,s cap, and a timeout otherwise. Every planner receives the same local goals from a shared $A^*$ global planner (0.10\,m grid, 5\,m lookahead) and the same active obstacle set, restricted to obstacles within 4.0\,m of the robot to emulate a finite perception range. After each solve completes, the simulation advances by the measured wall-clock solve time with the robot still tracking the previously committed trajectory, and only then does the new trajectory deploy. Slower planners therefore act on staler plans, as they would in deployment, while simulation overhead stays out of the measured time.

\subsection{Envelope and Model Ablations}\label{sec:exp-ablations}
Three ablations isolate the value of the fully-actuated $SE(2)$ model, with the full planner denoted \textbf{\texttt{holo}}. \textbf{\texttt{uni}} substitutes the unicycle model within the same framework. \textbf{\texttt{int-vx}} and \textbf{\texttt{int-vy}} replace the kinematics with a 3-DoF single integrator, and the velocity envelope \eqref{eq:ellipse} with the isotropic disc $v_x^2+v_y^2\le v_r^2$. An isotropic bound cannot match the real anisotropic bound on both axes, so each integrator ablation focuses on one. \texttt{int-vx} sets $v_r=3.0$\,m/s to match the true forward speed $v_{x,\max}$, overapproximating the envelope laterally at the cost of executability, as planned lateral velocities can exceed what the robot can execute. \texttt{int-vy} sets $v_r=1.5$\,m/s to match the lateral speed $v_{y,\max}$, a faithful but conservative subset of the envelope. 

Table~\ref{tab:results} reports results over 50 shared seeds, with solve time entering each planner's closed loop as described in Sec.~\ref{sec:exp-setup}. The conservative \texttt{int-vy} matches \texttt{holo} on static success and trails by one seed in the dynamic environment, but its shrunken envelope shows up in mission time, with \texttt{holo} reaching the goal roughly $1.7\times$ faster at comparable safety. The overapproximating \texttt{int-vx} instead loses successes, timing out on six static seeds while still trailing \texttt{holo} in mission time. Giving up the lateral degree of freedom, \texttt{uni} nearly matches \texttt{holo} on static success, but takes $\sim60\%$ longer, and falls further behind under moving traffic. Heading-coupled lateral motion buys time, and under moving obstacles, it buys some safety as well. Solve times are similar across the ablations, with only \texttt{uni} meaningfully faster, owing to one fewer control input.

Five of the dynamic seeds produced a collision for every planner tested, suggesting obstacle scenes that leave little room to escape within the velocity bounds. On the remaining seeds, \texttt{holo} collided on four, two of them grazes under 3\,N, all of which the more conservative \texttt{int-vy} cleared, so a smaller envelope buys safety in some configurations and pays in others. If obstacle trajectories were known, the certificate can be extended to their predicted motion (Sec.~\ref{sec:collision}), which may recover safety.

\subsection{External Baseline Comparison}\label{sec:exp-baselines}
While the ablations vary the kinematic model within our framework, the external baselines are complete planning approaches of their own. \textbf{\texttt{layered}} is the multistage CBF safety-filter architecture of \cite{yamaguchi2026layered}, which synthesizes a Poisson safety function from an occupancy map and enforces it through a predictive stage and a real-time CBF-QP stage on a single integrator model. We refresh the safety function and predictive stage at the published 15\,Hz perception rate and run the QP stage fixed at a best-case 200\,Hz. \textbf{\texttt{rmp}} is the locally reactive controller of Mattamala et al.~\cite{mattamala2022efficient}, run via its open-source release, which combines goal-attraction and collision-avoidance Riemannian motion policies over a local elevation map into an $SE(2)$ twist command, at its published 10\,Hz rate. So that each baseline performs as its authors intended, independent of implementation and hardware, we run them at their idealized or published rates, while our planner incurs its measured solve time as latency.

Every static failure of \texttt{layered} in Table~\ref{tab:results} is a timeout and every dynamic failure is a collision, while \texttt{rmp} completes half its static missions and collapses to 1/50 under moving obstacles, consistent with a purely reactive design. Eight of the eleven static timeouts of \texttt{layered} were genuinely stuck, as doubling the mission cap to 80\,s recovered only three additional successes. In the static condition, \texttt{holo} reaches the goal roughly $2.5\times$ and $2.7\times$ faster than \texttt{layered} and \texttt{rmp} respectively. \texttt{layered} fails through conservatism, timing out rather than colliding, while \texttt{rmp}'s purely reactive design only fails by colliding. \texttt{holo} shows neither failure mode in the static condition while sustaining the fewest collisions under motion.

\begin{table}[t]
\vspace{4pt}
\centering
\caption{Closed-loop results over $n{=}50$ shared seeds, comparing the
full planner against its kinematic ablations (indented) and external
baselines under static and dynamic obstacles. The external baselines
(\texttt{layered} and \texttt{rmp}) run at idealized fixed rates, so
solve time is not reported.}
\label{tab:results}
\setlength{\tabcolsep}{4pt}
\begin{tabular}{@{}lccccc@{}}
\toprule
planner & success & collision & timeout &
\makecell{solve\\med/p95 [ms]} & \makecell{mission\\{[s]}} \\
\midrule
\multicolumn{6}{@{}l}{\emph{static}} \\
\texttt{holo} (ours) & \textbf{50/50} & 0 & 0 & 5.2 / 8.8 & \textbf{9.7} \\
\quad\texttt{uni} & 49/50 & 0 & 1 & \textbf{3.2 / 6.4} & 15.5 \\
\quad\texttt{int-vx} & 44/50 & 0 & 6 & 5.1 / 9.4 & 15.1 \\
\quad\texttt{int-vy} & \textbf{50/50} & 0 & 0 & 4.1 / 8.4 & 16.5 \\
\cmidrule(l{2pt}){1-6}
\texttt{layered}~\cite{yamaguchi2026layered} & 39/50 & 0 & 11 & -- & 24.8 \\
\texttt{rmp}~\cite{mattamala2022efficient} & 25/50 & 25 & 0 & -- & 26.8 \\
\midrule
\multicolumn{6}{@{}l}{\emph{dynamic}} \\
\texttt{holo} (ours) & \textbf{41/50} & 9 & 0 & 5.5 / 9.3 & \textbf{11.6} \\
\quad\texttt{uni} & 38/50 & 12 & 0 & \textbf{3.0 / 6.4} & 16.2 \\
\quad\texttt{int-vx} & 38/50 & 12 & 0 & 5.3 / 9.4 & 14.0 \\
\quad\texttt{int-vy} & 40/50 & 10 & 0 & 4.3 / 8.5 & 18.6 \\
\cmidrule(l{2pt}){1-6}
\texttt{layered}~\cite{yamaguchi2026layered} & 23/50 & 27 & 0 & -- & 22.7 \\
\texttt{rmp}~\cite{mattamala2022efficient} & 1/50 & 49 & 0 & -- & 34.6 \\
\bottomrule
\end{tabular}
\end{table}

\subsection{Hardware Demonstration}\label{sec:system}
We deployed the full pipeline of Fig.~\ref{fig:overview}, the \texttt{holo} planner, the $A^*$ global layer, and a depth-camera perception pipeline, on a Boston Dynamics Spot with all computation on an onboard NVIDIA Jetson Thor. The pipeline fuses the point clouds of Spot's five body-mounted depth cameras into a single cloud, constructs an occupancy grid, clusters occupied cells, and fits a circle to each cluster, forming the obstacle set of \eqref{eq:collision}, refreshed at $\sim18$\,Hz. The planner then commands body velocities to Spot's black-box controller.

\begin{figure}[t]
  \centering
  \includegraphics[width=\columnwidth]{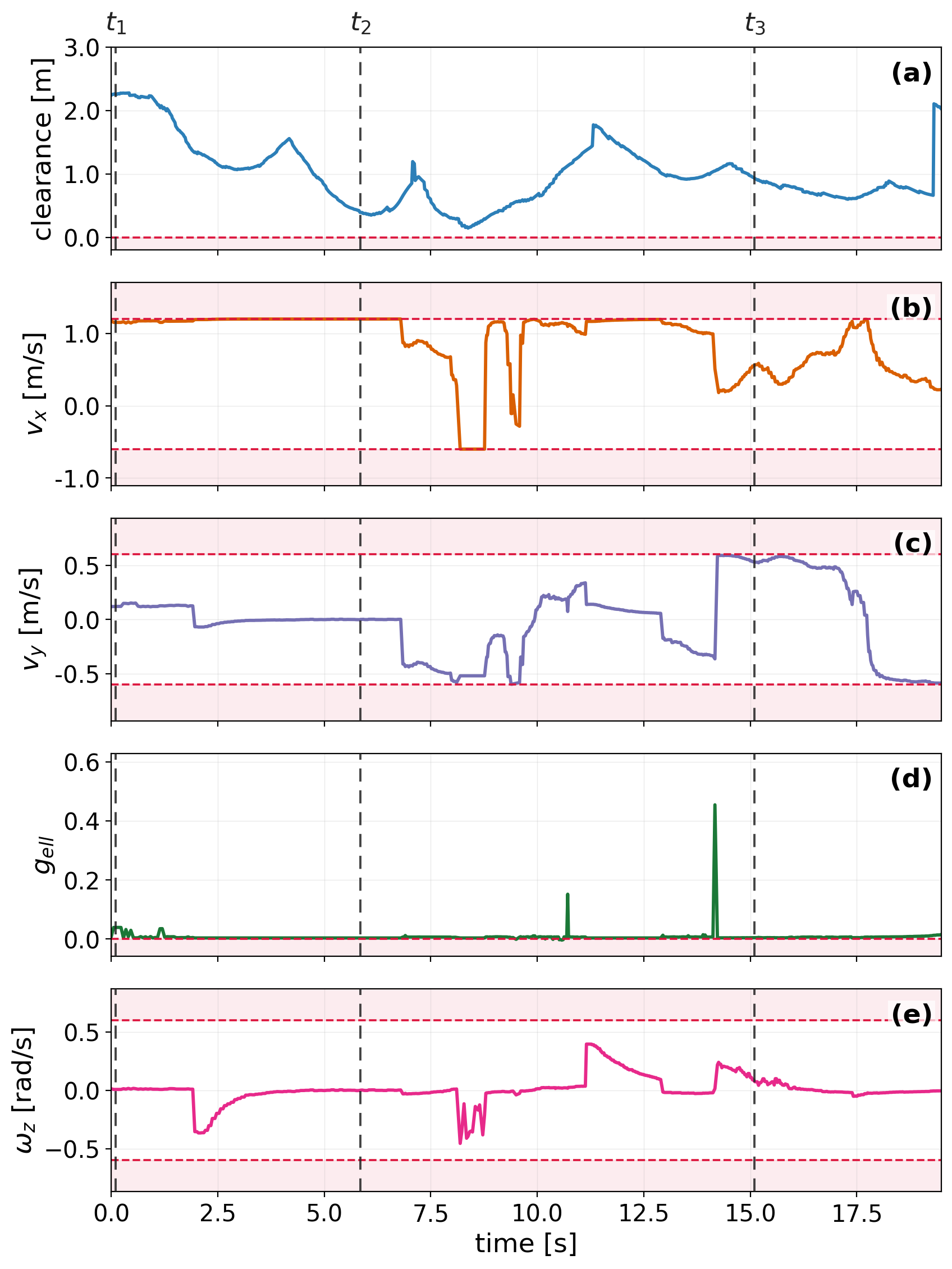}
  \caption{Constraint satisfaction over one corridor traversal (721 solves, 19.5\,s). (a)~clearance to the nearest perceived obstacle, (b)~commanded $v_x$, (c)~commanded $v_y$, (d)~coupled velocity-envelope margin $g_{\mathrm{ell}} = 1 - (v_x^2/1.2^2) - (v_y^2/0.6^2)$, (e)~commanded yaw rate. Clearance is measured from odometry. $t_1$--$t_3$ represent the stills from Fig. \ref{fig:hardware}.}
  \label{fig:pass-constraints}
\end{figure}

Twenty-two consecutive runs were conducted in which Spot crossed a field of five differential-drive robots (Fig.~\ref{fig:hardware}), with bounds $v_{x,\max}=1.2$\,m/s,
$v_{y,\max}=0.6$\,m/s, $\omega_{\max}=0.6$\,rad/s, and
$v_{x,\mathrm{rev}}=0.6$\,m/s, selected for outdoor operation among moving robots, with $W_p=1.0$ and $W_\omega=2.0$. Fig.~\ref{fig:pass-constraints} shows constraint satisfaction over a representative run, and the executed trajectories exercise coupled longitudinal--lateral motion, including lateral translation through gaps between the moving obstacles, as seen in Fig.~\ref{fig:hardware}. Across these 22 traversals totaling 392\,m, the planner produced 13,331 solves. Twenty traversals completed successfully, and two involved contact with an obstacle. The two contacts, like the four avoidable collisions in simulation, are likely attributable to the quasi-static assumption, where each solve certifies the planned trajectory against the obstacles as perceived at that instant, so obstacle motion within a replan cycle, error in perception, and tracking error beyond $\epsilon$ are not covered.

Over the full 22 traversals, the NLP solves in 16.3\,ms median (37.9\,ms p95) onboard, sustaining a 36\,Hz mean replan rate when considering overhead. The deployment shows the formulation operating under real perception, using obstacle estimates with real sensing noise and latency that is absent from the ground-truth simulation.

\section{Conclusion}
\label{sec:conclusion}
We presented a real-time motion planning framework that plans over the fully-actuated $SE(2)$ kinematic model of a quadruped and certifies velocity and collision constraints along the entire continuous trajectory. The approach accepts the nonconvexity that differential flatness is usually invoked to avoid, and the resulting NLP, transcribed by composite Bernstein polynomials, solves at replanning rates on onboard hardware. In closed-loop simulations our planner succeeded on all static trials, sustained the highest success rate under moving obstacles, and reached the goal $1.6$--$2.7\times$ faster than every ablation and external baseline in the static environment. On hardware, the full stack ran onboard at a 36\,Hz replan rate under depth-camera perception, executing coupled longitudinal--lateral motion among moving obstacles. Future work will address the residual gaps between the per-solve safety certificate and closed-loop safety, extending the certificates to obstacle predictions, and accounting for perception latency and tracking error.

\bibliographystyle{IEEEtran}
\bibliography{refs}

\end{document}